\pdfoutput=1

\documentclass[11pt]{article}

\usepackage[preprint]{acl}

\usepackage{times}
\usepackage{latexsym}
\usepackage{enumitem}

\usepackage[T1]{fontenc}

\usepackage[utf8]{inputenc}

\usepackage{microtype}
\usepackage{booktabs}
\usepackage{graphicx}
\usepackage{xspace}
\usepackage{amsfonts}
\usepackage{amsmath}
\usepackage{multirow}
\usepackage{listings}
\usepackage{xcolor}
\usepackage{xspace}
\usepackage{inconsolata}
\usepackage{verbatim}
\usepackage[font=small,skip=2pt]{caption}
\usepackage{pgfplots}
\pgfplotsset{compat=1.18}

\usepackage{url}

\hypersetup{breaklinks=true}

\newcommand\smallsection[1]{\noindent\textbf{#1.}}

\newcommand\system{\textsc{VeriLocc}\xspace}

\title{\system: End-to-End Cross-Architecture Register Allocation via LLM}

\author{
  Lesheng Jin$^1\quad$Zhenyuan Ruan$^2\quad$Haohui Mai$^3\quad$ Jingbo Shang$^{1}$ \\
  $^1$UC San Diego $\quad$ $^2$MIT $\quad$ $^3$CausalFlow Inc. \\
  $^1$\{l3jin, jshang\}@ucsd.edu $\quad$ $^2$zainruan@mit.edu $\quad$ $^3$haohui@causalflow.ai
}

\begin{document}
\maketitle

\begin{abstract}

Optimizing GPU compilers must find quality solutions of the combinatorial
compiler optimization problem (e.g., register allocations) to generate
performant GPU binaries. Currently they mostly rely on hand-crafted
heuristics which require substantial re-tuning for each hardware generation. 
We introduce \system, a framework that combines large language models (LLMs) with formal compiler techniques to enable generalizable and verifiable register allocation across GPU architectures. 
\system fine-tunes an LLM to translate intermediate representations (MIRs) into target-specific register assignments, aided by static analysis for cross-architecture normalization and generalization and a verifier-guided regeneration loop to ensure correctness. 
Evaluated on matrix multiplication (GEMM) and multi-head attention (MHA), \system achieves 85-99\% single-shot accuracy and near-100\% pass@100. 
Case study shows that \system discovers more performant assignments that outperform the state-of-the-art, expert-tuned rocBLAS by over 10\% in runtime.

\end{abstract}

\section{Introduction}
\label{s:intro}

\begin{figure*}[t]
\small\centering
\includegraphics[width = \linewidth]{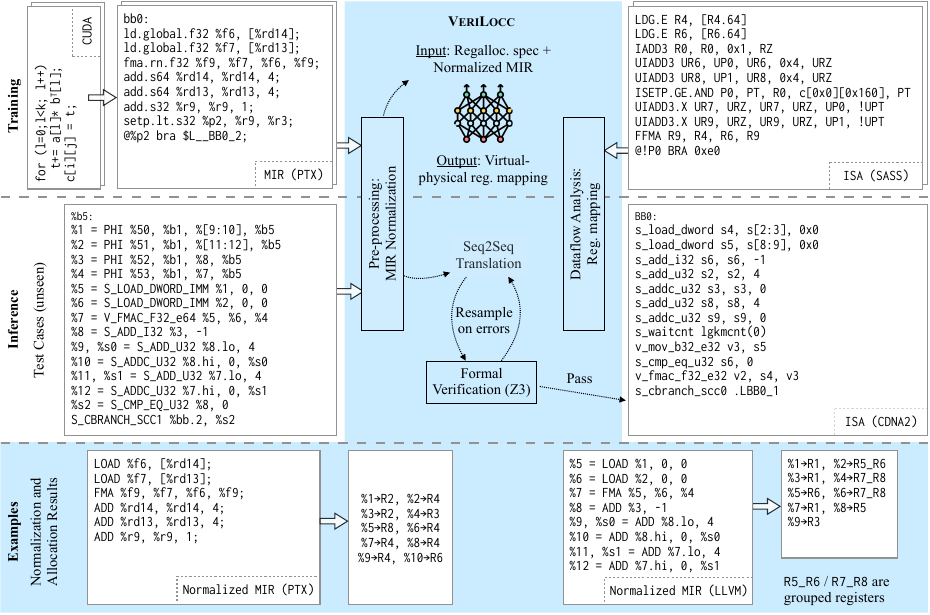}
\caption{Overall training and inference workflow of \system. 
Our seq2seq formulation learns to translate normalized MIR into virtual-to-physical register mappings, derived from MIR and ISA through dataflow analysis that tracks value propagation. 
For clarity, we show only the MIR (e.g., PTX) and ISA (e.g., SASS) fragments that load matrix \(A\) from memory and perform dot-product accumulation.
We also include illustrative examples of the normalized MIRs and the structured mapping between virtual and physical registers across architectures. 
}
\label{fig:ir_normalization}
\label{f:overview}
\end{figure*}

Modern GPUs have dramatically reshaped deep learning by offering massive parallel computational powers~\cite{krizhevsky2012imagenet}. 
Unleashing this performance requires not only hardware advances~\cite{Jouppi:2017,Wang:2021,Zhao:2025}, but also increasingly sophisticated software stacks~\cite{Chen:2018,LI:2023,Ma:2020,NVIDIA-Corporation:2024,Wu:2025,Zheng:2020}. 
At the core are optimizing compilers, which translate high-level GPU kernels into efficient, hardware-specific binaries.

Many of these optimizations are NP-complete; a central example is \emph{register allocation}~\cite{Alfred:2007}, as illustrated in Figure~\ref{f:overview}. 
The task involves assigning virtual registers in the compiler's intermediate representation (MIR) to physical registers in the instruction set architecture (ISA). 
A \emph{correct} allocation must (1) consistently map the same virtual register to the same physical register, and (2) assign virtual registers with overlapping lifetimes to disjoint physical registers.

Achieving \emph{performant} register allocation is even harder, as it requires
modeling architectural details such as register bank conflicts~\cite{Guan:2024}, pipeline stalls, and memory spill costs. 
Modern compilers rely on hand-crafted or learned heuristics~\cite{Lozano:2019,Quintao-Pereira:2008}, while performance-critical libraries such as BLAS~\cite{AMD-Inc.:2024} often resort to handwritten assembly optimized for specific GPUs. 
Both approaches demand substantial engineering effort and are difficult to retarget across hardware generations.

Recent work has explored learning-based allocations, but mainly rely on
handcrafted features~\cite{chen2021mlgo} or lack of the soundness guarantees to
integrate the model outputs into the real-world compiler
stacks~\cite{liu2024llmcompiler}. General-purpose models like ChatGPT have
limited abilities on NP-Complete problems~\cite{Wei:2025}. Therefore they often
struggle with register allocation, producing incorrect sequential assignments
that ignore liveness constraints (Appendix~\ref{sec:chatgpt_fails}). This
motivates our approach: combining the expressiveness of large language models
(LLMs)~\cite{openai2023gpt4,Qwen:2025} with formal verification techniques to
achieve correctness and cross-architecture generalization.

This paper introduces \system, a learning-based register allocator that formulates register assignment as an end-to-end sequence-to-sequence (seq2seq) translation task~\cite{sutskever2014sequence} task for LLM.
Our key insight is that while GPU architectures differ in low-level details, they share core traits, such as in-order pipelines and sharded register banks, which persist across vendors and generations. 
\system fine-tunes an LLM to translate MIR into target-specific register assignments, treating different MIRs as dialects of a shared computational language. 
Inspired by neural machine translation~\cite{sutskever2014sequence, bahdanau2014neural, vaswani2017attention}, \system learns to map virtual to physical registers while adapting to the syntactic and semantic variations of diverse GPU toolchains.

\system must overcome three key challenges to be practical. 
First, compiler transformations such as inlining and loop unrolling~\cite{Alfred:2007} can inflate MIR size beyond the LLM's effective context window~\cite{hsiehruler}; for instance, a single multi-head attention (MHA) kernel~\cite{vaswani2017attention} can exceed 50,000 tokens. 
Second, the model must generalize across architectures while respecting hardware-specific constraints like register file sizes and reserved registers. 
Third, correctness is critical -- an invalid allocation may silently corrupt a program or render it unexecutable.

\system addresses these challenges via a combination of normalization and
verification, as illustrated in Figure~\ref{f:overview}. \system employs static
analysis to normalize both MIR and ISA representations while preserving program
semantics~\cite{Lattner:2007,Xie:2007}. The analysis normalizes MIRs from
different compilers into a unified format, and extracts the results of register
allocations as JSON-style dictionaries (Figure~\ref{fig:ir_normalization}). This
reduces token length by 80–90\% in our experiments (Table~\ref{tab:dataset_stats}), improving LLM reasoning efficiency and allowing
the model to focus on allocation logic. Normalization also enables the creation
of a heterogeneous training dataset that combines compiler-generated outputs
from multiple toolchains with expert-optimized libraries—teaching the model both
general strategies and target-specific heuristics. To enforce correctness ,
\system models the generated mappings as instances of Satisfiability Modulo
Theories (SMT) problems, and formally verifies their correctness using
Z3~\cite{De-Moura:2008}. \system continues to re-sample for candidate allocations 
until a correct solution is found. 

We evaluate \system on two types of critical GPU kernels: general matrix
multiplication (GEMM) and multi-head attention (MHA), which together account for
over 90\% of inference time in modern LLMs~\cite{Dao:2022}. A fine-tuned 7B LLM achieves 85–99\%
single-shot correctness and near-100\% pass@100 with verification. A case study
on the GEMM kernel running on AMD’s MI250x GPU shows that \system discovers
register assignments that exploit architectural features overlooked by existing
compilers and human experts. The generated GEMM kernel outperforms
rocBLAS~\cite{AMD-Inc.:2024} by 11.6\% in runtime, which is the
state-of-the-art, hand-optimized BLAS library shipped by the GPU vendor. These
results demonstrate that \system combines the flexibility of data-driven
learning with the reliability required for deployment in production
compiler toolchains.

This paper makes the following contributions.
\begin{itemize}[nosep,leftmargin=*]
    \item We propose \system, a learning-based register allocator that combines LLMs, static analysis, and verifier-guided regeneration to achieve correctness and cross-architecture generalization.
    \item We evaluate \system on GEMM and MHA kernels, where a fine-tuned 7B LLM achieves 85–99\% single-shot accuracy and near-100\% pass@100, validating the feasibility of LLM-based register allocation.
    \item We show that \system can discover novel register assignments that outperform expert-tuned libraries, achieving 11.6\% runtime improvement over rocBLAS on AMD MI250x.
\end{itemize}

\smallsection{Reproducibility} Our implementation is publicly available on GitHub: \url{https://github.com/Jimmy-MMMM/VeriLocc}.

\section{Preliminaries: Register Allocations in GPU Compilers}

\smallsection{IR and MIR}
Modern compilers use Intermediate Representations (IRs) extensively for optimizations. 
The compilers take the application written in high level languages (e.g., CUDA) as inputs and then transform it to multiple levels of IRs. 
The compilers tackle different optimization goals at different levels of IRs. 
For example, LLVM~\cite{Lattner:2004} performs target independent optimizations such as constant propagations and dead code elimination at the level of LLVM IR, and performs target-specific optimizations such as coalescing memory accesses at the lower level machine IR (MIR). 
While different compiler toolchains have different naming (e.g., CUDA PTX vs SASS in the NVIDIA toolchains, and LLVM IR and LLVM MIR in the ROCm toolchain~\cite{amdROCm}), they share similar design principles. 
In the rest of the paper we use IR and MIR to refer to the target-independent and target-specific IRs.

\smallsection{Register Allocation}
Programs in MIR are not directly executable. 
MIRs use \emph{virtual registers}, or $\phi$-nodes in SSA forms~\cite{Bilardi:2003} to represent values in the programs. 
To realize the MIR to executable ISAs, the register allocators assign these virtual values to \emph{physical registers}. 
It must consistently assign the same value to the same register, and ensure that values with overlapping life cycles are assigned to disjoint registers. 
Additionally, the assignment must satisfy the hardware constraints (e.g., a 64-bit value must be assigned to two consecutive 32-bit registers). 
Due to the limited number of physical registers and hardware constraints, the register allocator might copy the values across registers, or temporarily spill the values to main memory. 

Finding performant results of register allocations is particularly important for GPU programs since it directly affects the available parallelism and instruction latency. 
A performant allocation would minimize the copies and spills, and consider low-level hardware features to minimize pipeline stalls. 
For example, NVIDIA GPUs organize the registers into 4 banks. 
Reading values in the same banks requires stalling the pipeline for another cycle, thus in Figure~\ref{f:overview} the compiler avoid the stalls by choosing \texttt{R4}, \texttt{R6}, and \texttt{R9} in the \texttt{FFMA} instruction which are in separate register banks. 
Note that finding the optimal register allocation is NP-complete~\cite{Chaitin:1981}, thus compilers often rely on either manual or learning-based heuristics when optimizing register allocations.

\smallsection{Complexity and Challenges}
Low-level code generation introduces substantial complexity that poses unique challenges for our sequence-to-sequence formulation. 
When compiled for GPUs, kernels undergo aggressive transformations such as inlining, loop unrolling, and memory coalescing. 
These optimizations significantly inflate the size of both the MIR and the final assembly. 
As shown in Table~\ref{tab:dataset_stats}, each multi-head attention (MHA) kernel averages 419 lines in PTX format, compared to 1,860 lines in LLVM IR. 
After tokenization, this corresponds to 9,294 tokens for PTX and 51,037 tokens for LLVM, far exceeding the context limits of current coding LLMs~\cite{hsiehruler}. 
As a result, naively fine-tuning LLMs struggles to capture long-range dependencies among register uses and assignments.

Architectural variability presents a second major challenge. 
Different GPU vendors employ distinct IRs and toolchains: NVIDIA uses PTX, while AMD uses a custom MIR format. 
As shown in Figure~\ref{fig:ir_normalization}, these representations differ in syntax, instruction structure, and register classes, making it difficult for cross-architecture modeling without careful normalization.

Finally, correctness is non-negotiable in compilers. 
A single invalid register assignment can lead to runtime crashes or silent corruption, rendering the compiled program unusable. 
Any learning-based allocator must therefore be paired with a mechanism that ensures compiler-level soundness.

These observations lead to three key challenges:
\begin{enumerate}[nosep,leftmargin=*]
    \item \textbf{Long sequences.} Optimized kernels produce IRs and ISAs that exceed the effective context window of typical LLMs, making long-range reasoning difficult.
    \item \textbf{Cross-architecture generalization.} The model must reconcile shared allocation strategies with ISA-specific syntax and constraints.
    \item \textbf{Compiler-level correctness.} The system must ensure sound register assignments to produce functional executables.
\end{enumerate}

\section{The \system Framework}

Figure~\ref{f:overview} illustrates the overall workflow of \system, which formulates register allocation as a sequence-to-sequence (seq2seq) transformation. 
Given tokenized MIR as input, the model generates structured register assignments in JSON format. 
However, raw MIR and ISA representations are often long, sparse, and inconsistent across architectures, making it difficult for LLMs to infer control flow and data dependencies implicitly. 
To address this, \system applies static analysis to normalize inputs and explicitly expose key semantic information at inference time. 
Finally, a verifier-guided regeneration loop ensures that the generated allocations satisfy semantic correctness.

\subsection{Seq2Seq Formulation}

Formally, given an input sequence $\mathcal{X}$ representing the tokenized MIR (potentially augmented with auxiliary information), the model generates an output sequence $\mathcal{Y}$ corresponding to register assignments in structured JSON format:
\begin{equation}
P(\mathcal{Y}|\mathcal{X}) = \prod_{i=1}^{m} P(y_i|\mathcal{X},
y_1, \ldots, y_{i-1})
\end{equation}
We train this mapping with a standard cross-entropy loss:
\begin{equation}
\mathcal{L} = -\frac{1}{m}\sum_{i=1}^{m}\log P(y_i|\mathcal{X},
y_1, \ldots, y_{i-1})
\end{equation}
\system normalizes the MIR before using it as input. 
The auxiliary information consists of the control and data dependency, the constraints of the task (e.g., the number of available registers), as well as hardware constraints on reserved registers and values. 
\system follows the control flow graph to allocate registers per basic block. 
Figure~\ref{fig:ir_normalization} presents two concrete examples of normalized MIRs along with their corresponding register mappings.

\subsection{Static Analysis-based Normalization}

\system performs static analysis for two tasks: 
(1) reconstructing the results of register allocations from the training data, as well as making control and data dependency explicit during inferences, 
and (2) normalizing the MIR / ISA to reduce the number of tokens and to improve generalizations.  

\subsubsection{Reconstructing Register Assignments} 

\system intercepts the open source ROCm toolchain to collect the register
assignments for AMD GPUs. However, the NVIDIA toolchain does not make the results of register allocations readily available.
To recover the mappings for training, \system transforms both the MIR and ISA to the Static Single Assignment (SSA) forms, and performs standard context-insensitive, path-insensitive, flow-sensitive global analysis~\cite{Xie:2007} to reconstruct the mappings. 
Particularly, \system follows the control flow graphs of both MIR and ISA, reconstructs the mappings by comparing the corresponding basic blocks, and finally consolidates the mappings for the full function. 
Using SSA forms allows \system easily to deal with the reordered instructions. 
\system uses heuristics to deal with cases where the toolchain chooses different instructions between MIRs and ISAs (e.g., lowering the \texttt{mul.wide.u32} instruction to a bit shift in Figure~\ref{fig:ir_normalization}).

\system uses the same techniques above to analyze the control flows and data dependency of the inputs. 
The information is later injected as auxiliary information into the input sequences.

\subsubsection{Normalizing MIRs} 

\system normalizes the MIRs for the training data and the inference inputs to reduce the number of tokens and to enhance generalizability across multiple architectures and workloads. 
First, it strips out irrelevant metadata (e.g., comments and debug symbols), and replaces the ISA-specific prologues (e.g., the pointer to the function arguments) as symbolic values since they are irrelevant to the task of register allocations. 
Second, it normalizes the instructions and register classes of different
architectures to common representations. For example, \system normalizes \texttt{fma.rn.f32} and \texttt{V\_FMAC\_F32\_e64} to \texttt{FMA} in the normalized MIR. 
\system also classifies a register to either a scalar or vector register, and
the type of the stored value (integer, float, or boolean). 
Additionally, \system renumbers the registers for each basic block to reduce the ranges of the tokens which improves the reasoning performance of the models.

\subsection{LLM Fine-tuning Pipeline}

\system uses a 7B decoder-only language model fine-tuned on the normalized
input-output pairs described above. As the backbone model, \system uses
\texttt{Qwen2.5-Coder-7B-Instruct} for its strong coding performance and
efficiency~\cite{Qwen:2025}. During training, the model is exposed to register
allocation examples from NVIDIA and/or AMD toolchains. Section~\ref{s:eval}
provides more details of the pipeline.

\subsection{Verifier-guided Inference}

At the inference time, the input is a normalized, tokenized MIR sequence along with auxiliary annotations, and the model generates a dictionary of register assignments in an auto regressive fashion. 
To eliminate hallucinations or incorrect results from the LLM, \system consists of a verifier to validate the results of register allocations before returning them to the compiler toolchain. 

\system constructs a SMT problem based on the input MIR to validate the followings:
\begin{itemize}[nosep,leftmargin=*]
\item {\em Consistency.} It consistently assigns the same virtual register to
the same physical register.
\item {\em Safety.} Virtual registers with overlapping life cycles are assigned
to disjoint physical registers. 
\item {\em Realizability.} The assignment satisfies the hardware constraints. 
\end{itemize}

Modeling the verification as SMT problems enables \system to reason about the solutions across various control flow paths, which is crucial for well-optimized performance-critical workloads like GEMM. 
Specifically, we use Z3 4.14.0 for validating the results of register allocations. 
It continuously re-samples the model until a valid allocation is found or a maximum number of attempts are exhausted.

\begin{table}[t]
    \centering
    \small
    \caption{Statistics of the data sets. The numbers of lines and tokens are averaged across different configurations. }
    \label{tab:dataset_stats}
    \setlength{\tabcolsep}{4pt} 
    \resizebox{\linewidth}{!}{
    \begin{tabular}{lcccc}
    \toprule
        \textbf{Kernel} & \textbf{\# Configs} & \textbf{GPU} & \textbf{\# Lines} & \textbf{\# Tokens} (raw / normalized) \\
    \midrule
        \multirow{2}{*}{GEMM} & \multirow{2}{*}{3,375}
        & NVIDIA & 79 & 1,233 / 113  $\quad$90.8\% $\downarrow$\\
        & & AMD & 71 & 2,289 / 274  $\quad$ 88.0\% $\downarrow$ \\
    \midrule
        \multirow{2}{*}{MHA} & \multirow{2}{*}{1,512}
        & NVIDIA & 419 & 9,294 / 1,944 $\quad$ 79.1\% $\downarrow$ \\
        & & AMD & 1,860 & 51,037 / 8,955 $\quad$ 82.5\% $\downarrow$ \\
    \bottomrule
    \end{tabular}
    }
\end{table}

\begin{table*}[t]
\caption{Effectiveness of \system in three different settings.
``w/o norm.'' is an ablation study of \system by disabling the MIR normalization.
It reports the numbers for both the GEMM and MHA test cases.
}
\label{tab:pass-rates}
\small
\resizebox{\linewidth}{!}{
\begin{tabular}{llccccc ccccc}
\toprule
& & \multicolumn{5}{c}{\textbf{GEMM} (768 test cases)} & \multicolumn{5}{c}{\textbf{MHA} (1865 test cases)} \\
\cmidrule(lr){3-7} \cmidrule(lr){8-12}
& & \multicolumn{2}{c}{\textbf{Greedy Decoding}} & \multicolumn{3}{c}{\textbf{Sampling Decoding}} & \multicolumn{2}{c}{\textbf{Greedy Decoding}} & \multicolumn{3}{c}{\textbf{Sampling Decoding}} \\
\cmidrule(lr){3-4} \cmidrule(lr){5-7} \cmidrule(lr){8-9} \cmidrule(lr){10-12}
\textbf{Setup} & \textbf{Model} & \textbf{Pass }$\uparrow$ / \textbf{Fail} $\downarrow$ & \textbf{Pass Rate} $\uparrow$ & \textbf{Pass@100} $\uparrow$ & \textbf{AvgTry} $\downarrow$ & \textbf{MaxTry} $\downarrow$ & \textbf{Pass }$\uparrow$ / \textbf{Fail} $\downarrow$ & \textbf{Pass Rate} $\uparrow$ & \textbf{Pass@100} $\uparrow$ & \textbf{AvgTry} $\downarrow$ & \textbf{MaxTry} $\downarrow$ \\
\midrule
\multirow{2}{*}{Same-NV} 
  & \system  &  764 / 4  &  \textbf{99.48\%}  &  \textbf{99.47\%} & \textbf{1.15} & \textbf{122}  & 1796 / 69 & \textbf{96.30\%}  & \textbf{99.74\%} & \textbf{1.44} & \textbf{149}  \\
  \cmidrule{2-12}
  & w/o norm. & 756 / 12  &  98.44\%  & 99.08\% & 1.51 & 158 & 1779 / 86 & 95.39\% & 99.74\% & 1.52  & 181  \\
\midrule
\multirow{2}{*}{Same-AMD} 
  & \system      & 764 / 4   & \textbf{99.48\%} & \textbf{99.86\%} & \textbf{1.10} & \textbf{183} & 1794 / 71  & \textbf{96.19\%} & \textbf{96.84\%} & \textbf{2.61} & \textbf{105}\\
  \cmidrule{2-12}
  & w/o norm.    & 762 / 6   & 99.22\% & 99.35\% & 1.15 & 201 & 1775 / 90  & 95.17\% & 95.87\% & 3.09 & 176 \\
\midrule
\multirow{2}{*}{Mixed} 
  & \system      & 753 / 15  & 98.05\% & \textbf{98.56\%} & 1.14 & \textbf{143} & 1601 / 264 & \textbf{85.84\%} & \textbf{89.76\%} & \textbf{6.37} & \textbf{227} \\
  \cmidrule{2-12}
  & w/o norm.    & 749 / 13  & 98.29\% & 98.05\% & 1.13 & 170 & 1491 / 374 & 79.95\% & 84.24\% & 9.11 & 241 \\
\bottomrule
\end{tabular}
}
\end{table*}

\section{Experiments}
\label{s:eval}

The evaluation aims to answer the following questions both qualitatively and
quantitatively:
\begin{itemize}[nosep,leftmargin=*]
\item How effective can \system generate correct register allocations?
\item How effective can \system generalize over new programs and
architectures?
\item How effective can normalizations improve model performance?
\item Is \system sufficiently fast to be used real-world settings? 
\end{itemize}

\subsection{Datasets} 

We curate data from two of the most computationally intensive kernels in
large-scale deep learning: general matrix multiplication (\textbf{GEMM}) and
multi-head attention (\textbf{MHA}). We collect the MIRs and ISAs of the GEMM
and MHA kernels of various configurations. 
Each configuration has its own shapes of memory tiles~\cite{Kjolstad:2017}, different levels of unrolling, and different lengths of software pipelines. 
The dataset closely resembles real-world libraries including BLAS and FlashAttention, which compute the results for different dimensions of inputs with GPU-specific configurations for best runtime performances. 
The dataset consists of MIRs and ISAs for both NVIDIA RTX 4090 and AMD MI250x GPUs. 
Table~\ref{tab:dataset_stats} describes the statistics of the data set.
The MHA kernels exhibit higher computational complexity with significantly more basic blocks per instance than GEMM, resulting in a total dataset of 9,325 instances.
Please refer to Appendix~\ref{sec:kernel_config} for more details.

\subsection{Evaluation setups}

We evaluate \system on two servers: one with equipped with 500GB SSD, 10GbE ethernet, and a RTX 4090 GPU.
The other server with equipped with 17TB SSD, 10GbE ethernet, and a MI250x GPU. Both servers run Ubuntu 22.04, CUDA 12.4 and ROCm 6.3.1.
We serve the models at the server with NVIDIA GPU and report the average of 100 runs of the runtime performances.  

We randomly partition the dataset and use 80\% / 20\% of the data for training and testing, respectively. 
We consider two experiment settings: 
(1) \textbf{Same-NV} and \textbf{Same-AMD}:
training and inferences are done with data on the same NVIDIA or AMD hardware architecture; and 
(2) \textbf{Mixed}, where training and inferences are done with data from both architectures.

Our primary metric is the \textbf{Pass Rate}, which measures the percentage of single-shot greedy decoding generations that pass the verification, i.e., produce a correct register allocation. 
We also report \textbf{Pass@100}, the proportion of test cases where at least one valid allocation is generated within 100 attempts. 
To assess runtime overhead if one integrates our LLM-based register allocator into the compiler toolchains, we additionally report the average attempts (\textbf{AvgTry}) and the maximum attempts (\textbf{MaxTry}) required to generate the first correct allocation.

\subsection{Main Results}

Table~\ref{tab:pass-rates} reports the number of passing and failing instances, along with the percentages of valid generations under one-shot (Pass@1) and up to 100 attempts (Pass@100). 
It also includes the average and maximum number of attempts required to produce the first valid allocation. 

\system achieves strong single-shot performance, with Pass@1 rates ranging from 85\% to 99\% across GEMM and MHA test cases. 
MHA kernels pose greater difficulty due to longer sequences and more complex dependencies, with a Pass Rate of 85.84\% in the mixed training setting. 
Nevertheless, a simple resampling strategy proves highly effective, Pass@100 is nearly 100\%, with the average number of attempts under 10, and maximum attempts typically falling within 100 to 200.

Overall, \system works very well in the both the Same-NV / Same-AMD settings.
While the Mixed setting is more challenging, \system still generalizes well. 
This indicates that LLMs can learn transferable patterns in register allocation across different ISAs.

\subsection{Ablation on MIR Normalization} 
The MIR normalization is effective at reducing the input token numbers by about 80\% to 90\% as shown in Table~\ref{tab:dataset_stats}. 
We further compare the end-to-end effect of disabling the normalization in Table~\ref{tab:pass-rates}.
The normalized version consistently perform better in terms of almost all evaluation metrics. 
This confirms the importance of our proposed MIR normalization.

\begin{figure}[t]
    \centering
    \input{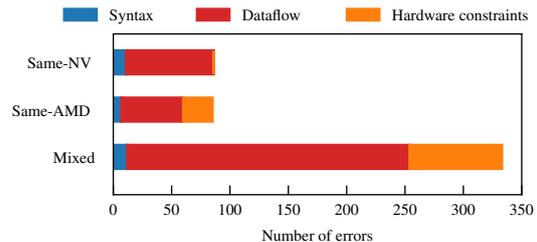}
    \caption{Error Distributions of \system on MHA under Different Settings. }
    \label{fig:error_distribution}
\end{figure}

\subsection{Error Analysis of Greedy Decoding}

We analyze all error cases from \system's greedy decoding on MHA to better understand its limitations. 
GEMM is excluded because the total number of errors on GEMM is too small.
Figure~\ref{fig:error_distribution} shows the distribution of error types across different settings. 
Errors fall into three main categories
\begin{itemize}[nosep,leftmargin=*]
    \item \textbf{Syntactic Errors}: 
        These account for 4--8\% of failures and involve malformed JSON output, e.g., missing braces, broken key-value pairs, or extraneous text. 
        Such errors are typically easy to fix via post-processing.

    \item \textbf{Dataflow Violations}: 
        The most severe, comprising 62--73\% of errors, where \system overwrites values that are still live later in the program. 
        These semantic violations directly compromise program correctness.

    \item \textbf{Hardware Constraint Violations}: 
        Making up 24--29\% of errors, these occur when the model assigns values to register configurations that violate hardware rules, for example, using non-consecutive registers (\texttt{vgpr0\_vgpr2}) for 64-bit operands, where contiguous pairs (e.g., \texttt{vgpr0\_vgpr1}) are expected.
\end{itemize}
Interestingly, the proportion of dataflow violations is lower when the model is trained on a single architecture, suggesting that cross-architecture generalization remains a key challenge despite our normalization efforts.
As a future direction, one could incorporate constrained decoding that explicitly masks out currently occupied registers during generation, helping to reduce both dataflow violations and hardware constraint errors.

\subsection{Efficiency in End-to-End Compilation} 
The NVIDIA and AMD toolchains compile the test cases in 1107 ms on average. 
\system spends most of the time in two components: model servings and verification. 
Our evaluation server serves the Qwen2.5-Coder-7B-Instruct model at the speed of 8260 input tokens per second and 131 output tokens per second with no concurrent requests. 
Our end-to-end measurements closely match the numbers: the average time of serving a single inference in \system is 945 ms. 
\system can adopt techniques like continuous batching~\cite{Kwon:2023} and radix attentions~\cite{Zheng:2024} to significantly speed up the
inferences. 
This is left to future work. 

The verifications can take significant more time. 
On average it takes 31.77 ms to verify the correctness of the assignments with Z3. 
The longest verification takes 7.7 seconds. 
While validating worst case assignments does require the full capabilities of Z3, it is possible to accelerate the common cases where the verification can be done via following the control and data
dependency in polynomial time. 
We leave this optimization to future work.

\subsection{Case Study: Beyond Existing Compilers}

One thing worth noting is that \system is able to find a more performant solution for GEMM compared to rocBLAS, the state-of-the-art GEMM library offered by AMD. 
This refined version of GEMM achieves 111.44 TFLOPS when multiplying $128 \times 4096$ and $4096 \times 4096$ \texttt{fp16} matrices with a strided batch of $3$, which is 11.63\% faster than rocBLAS. 
A detailed analysis shows that \system discovers a more performant assignment than the one used in the expert-optimized rocBLAS. 
Particularly, the MI250x GPU is based on the CDNA2 architecture, which introduces dedicated matrix core units to accelerate matrix multiplications. 
It offers two types of vector registers: standard architectural vector registers (VGPRs) or accumulation VGPRs (AccVGPRs), where the VGPRs can be used by all compute components but the AccVGPRs are exclusive to the matrix core units. 
\system discovers an assignment that stores the values of the matrices in the AccVGPRs aggressively while rocBLAS stores them in the VGPRs.
The publicly available  ISA documentation~\cite{mi200isa} indicates there should be no performance differences. 
We suspect that the matrix core units have faster access to the AccVGPRs at the micro architecture level, which has a lower latency thus improves the performance. 
Though anecdote, it demonstrates the practical advantage and potential of \system over expert-optimized libraries.

\subsection{Performance of 3B Model}

We evaluate \texttt{Qwen2.5-Coder-3B-Instruct} to assess the feasibility of using smaller models for register allocation. 
While the 3B model achieves reasonable performance on GEMM (up to 98\% pass@100 with sampling), it consistently fails on MHA tasks. 
This highlights a clear capability gap: smaller models struggle with the complex dependencies and register pressure present in real-world kernels like MHA. 
Notably, our normalization method continues to provide benefits at this scale, improving pass rates and reducing decoding attempts. 
These results suggest that more sophisticated compiler tasks may require larger models to maintain effectiveness.

\begin{table}[t]
\caption{Effectiveness of \system while using 3B model on GEMM. MHA is too hard for 3B model.}
\label{tab:smaller-models}
\small
\resizebox{\linewidth}{!}{
\begin{tabular}{llcccc}
\toprule
& & \multicolumn{1}{c}{\textbf{Greedy Decoding}} & \multicolumn{3}{c}{\textbf{Sampling Decoding}} \\
\cmidrule(lr){3-3} \cmidrule(lr){4-6} 
\textbf{Setup} & \textbf{Model} & \textbf{Pass Rate} $\uparrow$ & \textbf{Pass@100} $\uparrow$ & \textbf{AvgTry} $\downarrow$ & \textbf{MaxTry} $\downarrow$ \\
\midrule
\multirow{2}{*}{Same-NV} 
  & \system   &  \textbf{89.06\%}  &  \textbf{98.44\%} & \textbf{6.84} & \textbf{782}    \\
  \cmidrule{2-6}
  & w/o norm.   &  88.02\%  & 97.14\% & 7.03 & 977   \\
\midrule
\multirow{2}{*}{Same-AMD} 
  & \system        & \textbf{87.24\%} & \textbf{98.67\%} & \textbf{7.29} & \textbf{793} \\
  \cmidrule{2-6}
  & w/o norm.      & 86.46\% & 98.67\% & 7.85 & 993 \\
\bottomrule
\end{tabular}
}
\end{table}

\section{Related Work}

\smallsection{Learning-based Compiler Optimization}
Recent advances in machine learning have opened new directions for compiler
optimization by replacing handcrafted heuristics with learned models.
MLGO~\cite{chen2021mlgo} applies reinforcement learning in LLVM for tasks like
inlining and register allocation, but depends on manual features and tight
coupling with compiler internals. Meta's LLM Compiler~\cite{liu2024llmcompiler}
trains LLMs on IR and assembly for end-to-end translation. The generated
results, however, lack the essential soundness guarantees to be
consumed in the compiler stacks. General-purpose models like ChatGPT also
struggle with register allocation, producing sequential, invalid assignments
without liveness reasoning (Appendix~\ref{sec:chatgpt_fails}). In contrast, our
work combines the expressiveness of LLMs with formal verification techniques,
using normalization and verifier-guided decoding to ensure correctness and
cross-architecture generalization.

\smallsection{Optimizing tensor compilers}
Modern AI models are essentially
tensor programs. Optimizing tensor
compilers~\cite{Chen:2018,LI:2023,Ma:2020,Ragan-Kelley:2013,Wu:2025,Zheng:2020}
realize tensor programs to performant implementations on GPU or ASICs~\cite{Jouppi:2017}.
They automatically apply techniques such as memory tiling~\cite{Ragan-Kelley:2013}, operator
fusions~\cite{Chen:2018}, software pipelining~\cite{Ma:2020} to search efficient schedules to
utilize the hardware. They mostly operate at the tensor level on intermediate
representations such as MLIR~\cite{Lattner:2021}, and delegates the effort of generating
low-level ISAs to the GPU / ASIC toolchains. \system operates at a lower level
of the software stack. It focuses on improving the register allocations inside
the toolchains. They can be combined together to further improve performances.

\smallsection{Register allocation} 
Register allocation is one of the integral components of the compiler backends.
Optimal register allocation is NP-Complete, therefore compilers uses manual
heuristics, constraint solvers~\cite{Quintao-Pereira:2008}, combinatorial
optimizations~\cite{Lozano:2019} and reinforcement
learnings~\cite{venkatakeerthy2023rl4real} to find satisfactory solutions.
Compilers also opt for linear register allocations~\cite{Poletto:1999} in
latency-sensitive use cases. \system leverages LLMs to {\em learn} effective
strategies of register allocations directly from the MIRs and ISAs, opening a
new path path toward generalizable, semantics-aware compiler optimization.

\smallsection{Static analysis and validations} 
Static analysis~\cite{Lattner:2007,Xie:2007} is effective to detect errors and
vulnerabilities in the programs. \system uses sound static analysis~\cite{Mai:2023}
(i.e., no false negative) to effectively validate the results of register
allocations are correct.

\section{Conclusions and Future Work}

We demonstrate that large language models (LLMs) can effectively learn register allocation as a sequence-to-sequence task, achieving high correctness rates and competitive runtime performance. 
\system generalizes across architectures, benefits from normalization, and even surpasses expert-optimized libraries in some scenarios, highlighting the potential of learning-based approaches in compiler backends.

A key direction for future work is integrating the verifier into training.
One possibility is to provide verifier feedback as additional input to the model during fine-tuning. 
Another avenue is to explore reinforcement learning, where the verifier acts as a reward signal to directly guide the model toward sound and performant allocations.

\section*{Limitations}

While \system shows strong results on register allocation for key GPU kernels, it has several limitations. 
First, our evaluation focuses on structured compute kernels (GEMM and MHA) that dominate LLM inference workloads; generalizing to less regular or control-heavy programs remains future work. 
Second, although we demonstrate cross-architecture generalization between NVIDIA
and AMD backends, adaptation to vastly different architectures (e.g.,
CPU and TPU) would require additional normalization and fine-tuning. 
Third, the model's performance degrades for highly complex kernels when using smaller architectures (e.g., 3B), suggesting that larger models may still be necessary for difficult compiler tasks. 
Finally, while our verifier ensures correctness, the re-sampling loop can add latency in low-confidence cases.

We believe these are meaningful directions for future work, including scaling to more optimization targets, improving efficiency, and expanding to broader compiler infrastructures.

\bibliography{anthology,custom}

\appendix
\section{Kernel Configurations}
\label{sec:kernel_config}

A GEMM kernel calculates $C = \alpha \cdot A \times B + \beta \cdot C$ with matrices $A \in \mathbb{R}^{M \times K}$, $B \in \mathbb{R}^{K \times N}$, and $C \in \mathbb{R}^{M \times N}$. 
To comprehensively evaluate register allocation strategies, we vary the matrix dimensions $M$, $N$, and $K$ across a range of values from 2 to 16, resulting in $15 \times 15 \times 15 = 3,375$ distinct problem configurations. 
This breadth of configurations allows us to observe how register allocation patterns adapt to different memory access patterns and computational requirements.

MHA is the defining component of Transformer architectures, consisted of several computational phases: (1) projection of queries, keys, and values through linear transformations, (2) computation of attention scores through scaled dot-product attention, and (3) application of attention weights to values, followed by a final linear projection. 
This complex sequence of operations introduces intricate dependencies and memory access patterns, making optimal register allocation particularly challenging and impactful for performance.

We sample a large configuration space inspired by real LLM inference workloads:
\begin{itemize}[nosep,leftmargin=*]
    \item Heads: [16, 24, 32, 40, 64, 128]
    \item Batch Size: [1, 32, 64, 128, 256, 512, 1024]
    \item Head Dim: [32, 40, 59, 64, 80, 96, 111, 128, 160, 192, 224, 256]
    \item Sequence Length: [2048]
    \item Attention Group: [1, 4, 8]
\end{itemize}
This parameter space yields $6 \times 7 \times 12 \times 1 \times 3 = 1,512$ distinct configurations, representing a comprehensive sampling of the operational demands placed on modern LLM inference systems.

\newpage

\section{ChatGPT Results}
\label{sec:chatgpt_fails}

\begin{figure}[h]
    \centering
    \includegraphics[width=\linewidth]{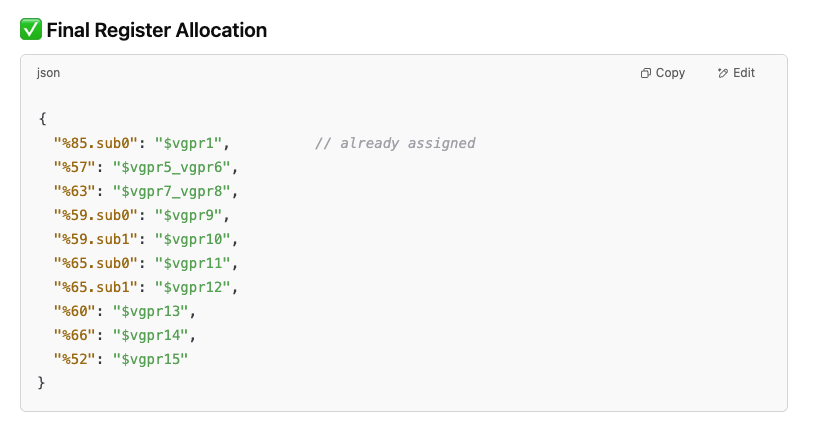}
    \caption{Allocation Result from ChatGPT}
    \label{fig:chatgpt}
\end{figure}

We analyze ChatGPT's behavior\footnote{\url{https://chatgpt.com/share/682a07f9-bf50-8006-a7ac-f07b439439e4}} on the register allocation task and observe substantial limitations in its ability to handle this core compiler optimization. 
As shown in Figure~\ref{fig:chatgpt}, ChatGPT outputs a naive sequential assignment strategy that fails to employ register reuse. 
This is reflected in the monotonically increasing register identifiers (\texttt{vgpr1} through \texttt{vgpr15}), with no attempt to recycle registers whose associated variables have expired. 
The result suggests a lack of liveness-aware reasoning and highlights the need for more specialized modeling or structural supervision to support compiler-level correctness.

\end{document}